\documentclass[letterpaper,10pt]{article}
\usepackage{authblk}
\usepackage{osameet3} %% use version 3 for proper copyright statement
\usepackage{amsmath,amssymb}

\usepackage[colorlinks=true,bookmarks=false,citecolor=blue,urlcolor=blue]{hyperref} %pdflatex

\usepackage{graphicx,amsmath,amsfonts,amsthm,amscd,amssymb,bm,url,color,latexsym}
\usepackage{physics}
\allowdisplaybreaks
\usepackage[capitalize,nameinlink]{cleveref}

\renewcommand{\mathbf}{\boldsymbol}

\newcommand{\mb}{\mathbf}
\newcommand{\mc}{\mathcal}

\newcommand{\bb}{\mathbb}

\newcommand{\reals}{\bb R}

\newcommand{\R}{\reals}
\newcommand{\Cp}{\bb C}

\newcommand{\paren}{\pqty}

\makeatletter
\newcommand{\printfnsymbol}[1]{%
  \textsuperscript{\@fnsymbol{#1}}%
}
\makeatother

\begin{document}

% \title{Fourier Phase retrieval  with a single measurement using Deep Learning}

% \title{Unlocking symmetry using structural models: An application in Fourier Phase Retrieval}
\title{Phase Retrieval using Single-Instance Deep Generative Prior}

\author{Kshitij Tayal\footnotemark[1*] \qquad Raunak Manekar\footnotemark[1] \qquad  Zhong Zhuang \footnotemark[2] \qquad David Yang \footnotemark[3] \qquad Vipin Kumar \footnotemark[1] \qquad Felix Hofmann \footnotemark[3]
\qquad Ju Sun \footnotemark[1*]}
\address{$^1$Department of Computer Science and Engineering, University of Minnesota, Twin Cities, USA \\ $^2$ Department of Electrical and Computer Engineering, University of Minnesota, Twin Cities, USA \\ $^3$ Department of Engineering Science, University of Oxford, Oxford, UK}
\email{ *\{tayal, jusun\} @umn.edu}

% \author{Kshitij Tayal,\authormark{$^1$*} Raunak Manekar,\authormark{$^1$}  Zhong Zhuang,\authormark{$^2$} David Yang,\authormark{$^3$} Vipin Kumar,\authormark{$^1$} Felix Hofmann,\authormark{$^3$} and Ju Sun\authormark{$^1$*} }

% \address{\authormark{$^1$} Department of Computer Science and Engineering, University of Minnesota, Twin Cities, USA\\
% \authormark{$^2$} Department of Electrical and Computer Engineering, University of Minnesota, Twin Cities, USA\\
% \authormark{$^3$} Department of Engineering Science, University of Oxford, UK}

% \email{\authormark{*}(tayal,jusun)@umn.edu} %% email address is required

%% Uncomment the following line to override copyright year from the default current year.
\copyrightyear{2021}

\begin{abstract}
Several deep learning methods for phase retrieval exist, but most of them fail on realistic data without precise support information. We propose a novel method based on single-instance deep generative prior that works well on complex-valued crystal data.
% Most neural networks for solving phase retrieval (PR) employ supervised training and need extensive training datasets to optimize their weights. Despite massive training datasets, deep networks fail to work due to the intrinsic symmetries. In this work, we propose using deep structural models that can accurately solve PR problems without any training datasets. To the best of our knowledge, this is the first work to consider structural models for Fourier phase retrieval.
\end{abstract}

\vspace{2em}
\section{Introduction}

Phase retrieval (PR) is a classical nonlinear inverse problem in computational imaging. The problem concerns recovering a complex signal $\mb X \in \Cp^{n \times n}$ from the oversampled Fourier magnitudes $\mb Y = \abs{\mc F\paren{\mb X}}^2 \in \R^{m \times m}$, where $m \ge 2n-1$ is necessary for recoverability. The problem has three intrinsic symmetries: any of 1) 2D translation of the nonzero content of $\mb X$, 2) 2D conjugate flipping of $\mb X$,  and 3) global phase offset to $\mb X$ ($\mb X e^{i \theta}$ for any $\theta \in [-\pi, \pi)$) and their compositions will leave the observation $\mb Y$ unchanged.

When $\mb X$ is real-valued and positive, numerous classical methods such as hybrid input-output (HIO, \cite{fienup1982phase})  can take advantage of the realness and/or positivity constraints to recover $\mb X$ in practice. However, when $\mb X$ is complex-valued---which pertains to most real applications, these constraints are not applicable. In such scenarios, knowing the precise support of $\mb X$ proves crucial for empirical successes~\cite{marchesini2003x}. Practical iterative algorithms for PR typically start with a loose support estimated from the autocorrelation (as $\mc F^{-1}(\abs{\mc F(\mb X)}^2)$ leads to the autocorrelation $\mb X \star \mb X$, from which one can obtain a crude estimate of the support of $\mb X$---less reliable when $\mb Y$ is affected by noise), and then gradually refine the support using thresholding (e.g., the popular shrinkwrap heuristic~\cite{marchesini2003x}) as the iteration proceeds. But the final recovery quality is often sensitive to the parameter setting in support refinement.

Recently, two families of deep learning (DL) methods have been proposed for PR. They either directly learn the inverse mapping parametrized by deep neural networks (DNNs) based on an extensive training set, or refine the results from classical methods by integrating DL modules with classical iterative algorithms. But,  as discussed in our prior work~\cite{manekarnips,TayalEtAl2020Unlocking}, most of these methods are only evaluated on real-valued natural image datasets that are distant from PR applications and also do not reflect the essential difficulty of PR. Here, we focus on complex-valued PR in real applications.

We consider the single-instance setting for PR. The most natural formulation used is probably the least squares
\begin{align}
\min_{\mb X \in \Cp^{n \times n}} \; \|\mb Y - \abs{\mc F\paren{\mb X}}^2\|_{F}^2.
\end{align}
Empirically, this almost always fails to recover anything meaningful on even simple test cases, probably due to the prevalence of bad local minimizers. Here, we propose a simple modification to make it work: parameterizing $\mb X$ using a deep generative prior, i.e., replacing $\mb X$ by $G_\theta\paren{\mb z}$, where $G_\theta$ is a trainable DNN parameterized by $\theta$, and $\mb z$ is a fixed seed vector. This leads to our main formulation for PR:
\begin{align}
\min_{\mb \theta} \; \|\mb Y - \abs{\mc F \circ G_\theta\paren{\mb z}}^2\|_{F}^2.
\end{align}

Recently, deep generative priors of the above form $G_\theta\paren{\mb z}$ have been used to solve inverse problems in computational imaging and beyond~\cite{ongie2020deep}. There are two major variants: 1) $G_\theta$ as a generator is pretrained on a large training set using deep generative models, e.g., GANs, and $\mb z$ is trainable. We call this variant \emph{multiple-instance deep generative prior} (MIDGP). The training set needs to come from the same domain as the object to be recovered; 2) $G_\theta$ is trainable and $\mb z$ is either fixed or trainable. We call this variant \emph{single-instance deep generative prior} (SIDGP), also referred as the untrained DNN prior. Deep image prior~\cite{deepimageprior} and deep decoder~\cite{heckel2018deep} are two notable models of SIDGP, and they only differ in the choice of architecture for $G$. Here, we take the SIDGP approach, as it does not need a training set, which may be expensive to collect for certain applications of PR.

SIDGP has been proposed to solve several variants of PR, including Gaussian PR~\cite{jagatap2019algorithmic}, Fourier holography~\cite{lawrence2020phase}, and phase microscopy~\cite{bostan2020deep} that all simplify PR to certain degrees. But no claim on PR has been made before the current work.

\section{Experimental Results}

\begin{figure}[!htbp]
  \centering
  \begin{minipage}[b]{0.4\textwidth}
    \centering
    \includegraphics[width=\textwidth]{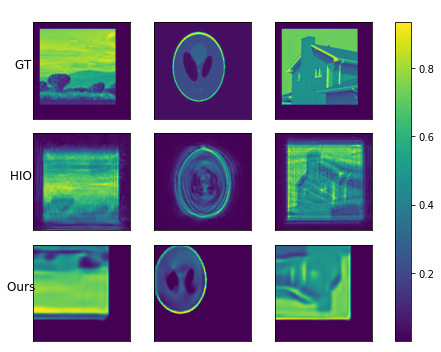}
    % \caption{
  \end{minipage}
  \begin{minipage}[b]{0.49\textwidth}
    \includegraphics[width=\textwidth]{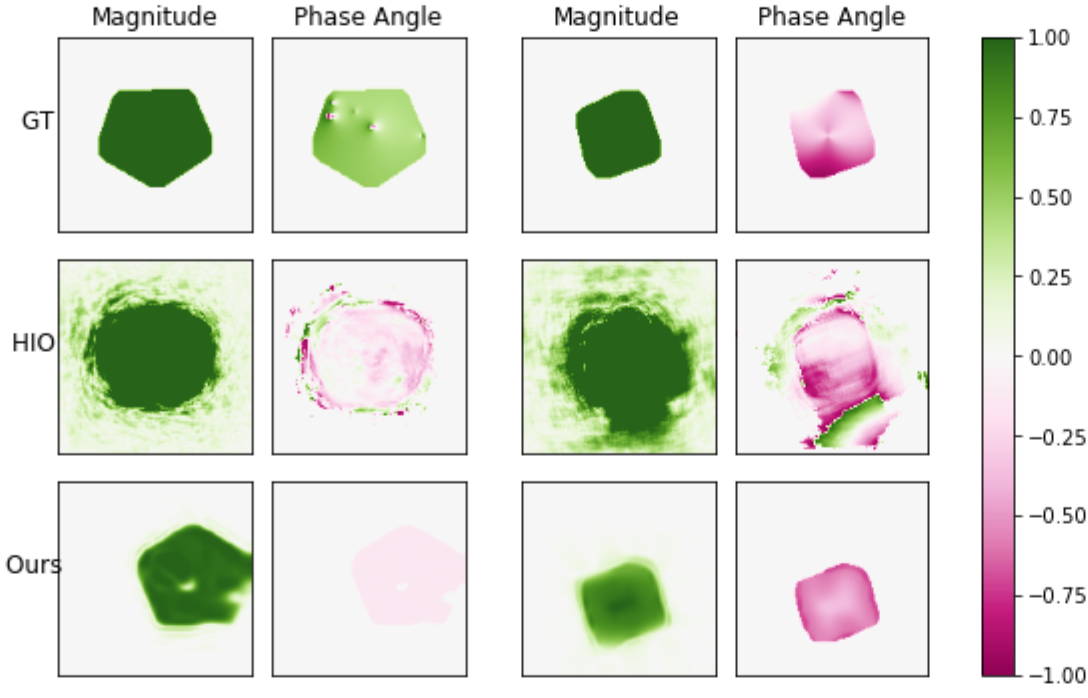}
    % \caption{Visualization of recovery results. First row contains complex groundtruth images, and the 2 nd and 3 rd columns are reconstructions produced by HIO and Structural Models, respectively.
  \end{minipage}
   \caption{Visualization of recovery results on (left) real-valued random toy images and (right) complex-valued simulated crystal data. For each of them, the first row is the groundtruth, and second row is the recovered result by HIO, and the third row by our method. }
     \label{fig:result}
\end{figure}

We test our method on several real-valued toy images and simulated complex-valued crystal data. Testing on the crystal data is inspired by the Bragg CDI application for material study~\cite{robinson2001reconstruction}. The data is generated by first creating 2D convex and concave shapes based on random scattering points in a $110\times 110$ grid on a $128 \times 128$ background. The complex magnitudes are uniformly $1$, and the complex phases are determined by projecting the simulated 2D displacement fields (due to crystal defects) to the corresponding momentum transfer vectors for each reflection. As shown in \cref{fig:result},
we do not assume the knowledge of the precise support, and hence the objects have translation freedom on all test images.

In all cases, our method produces good visual recovery results, while HIO, a representative classical algorithm for PR, leads to much worse recovery on the real-valued toy images, and completely fails on the complex-valued crystal data. We have used the plain version of HIO. Although incorporating support refinement strategies such as shrinkwrap will likely improve the results of HIO, it is amazing that our method based on SIDGP does not need any special handling of the support and works reasonably well.

% In this section, we empirically show that structural models are able to accurately recover the object upto its intrinsic symmetry. We use natural images with translation (real valued) and simulated crystal dataset(complex imaged) to test our method. We compared our method with popular iterative method HIO. Results in Figure \ref{fig:result}. Visibly, we can observe that HIO reconstruction has blurred boundary because of the translation and symmetry. We note that structural models give good recovery for both complex and real images and is able to resolve the symmetry without any difficulty.
% While this work demonstrated our approach success on simulated observations, we expect to conform its finding on experimental data in future work.

% We acknowledge that sometimes structural models doesn not converge and in that case we rerun our model with a different z.

% \input{problem_setup}
% \input{results}

\bibliographystyle{IEEEtran}
\bibliography{main}

\end{document}